# Application of Fuzzy Assessing for Reliability Decision Making

Shoele Jamali, and Mehrdad J. Bani, *Member, IAENG*

*Abstract*—This paper proposes a new fuzzy assessing procedure with application in management decision making. The proposed fuzzy approach build the membership functions for system characteristics of a standby repairable system. This method is used to extract a family of conventional crisp intervals from the fuzzy repairable system for the desired system characteristics. This can be determined with a set of nonlinear parametric programing using the membership functions. When system characteristics are governed by the membership functions, more information is provided for use by management, and because the redundant system is extended to the fuzzy environment, general repairable systems are represented more accurately and the analytic results are more useful for designers and practitioners. Also beside standby, active redundancy systems are used in many cases so this article has many practical instances. Different from other studies, our model provides, a good estimated value based on uncertain environments, a comparison discussion of using fuzzy theory and conventional method and also a comparison between parallel (active redundancy) and series system in fuzzy world when we have standby redundancy. When the membership function intervals cannot be inverted explicitly, system management or designers can specify the system characteristics of interest, perform numerical calculations, examine the corresponding α-cuts, and use this information to develop or improve system processes.

## I. INTRODUCTION

THE component redundancy plays a key role in engineering design and can be effectively used to increase system performances [1]. There are two common types of redundancy that are used, namely active redundancy, which stochastically leads to consideration of maximum of random variables, and standby redundancy, which stochastically leads to consideration of the convolution of random variables. Among studies considering imperfect coverage, [2] examined a model of a high voltage system consisting of a power supply and two transmitters with imperfect coverage in which the failure rate of fault coverage is constant. Recently, [3] proposed a reliability model with three phases of failure handling: failure detection, location, and recovery for continued service. Reported research has largely been concerned with obtaining measures of system effectiveness. However, while these results can be useful in analysis and assist the decision process, there are very few studies that establish a decision model which directly determines an optimum maintenance strategy. In [4] constructed a decision model using a Bayesian approach and selected utility functions. Their approach motivates us to develop an alternative method to analyze repairable systems in which the uncertainty of the parameters is accounted for using a fuzzy approach. Repairable systems formulated in this way have broader applications for reliability engineers and management than conventional models. A number of authors have investigated the two-unit redundant systems under different assumptions [4], [5], [6], [7], [8], and [9]. Almost all of the researches consider perfect coverage for failure, which means detection and recovery from failure has been successful. In practice, however, it may be impossible to replace the failed unit with a spare and then recover from a failure which means imperfect coverage [10].

In the literature described above, times to failure and times to repair are required to follow certain (known) probability distributions with fixed parameters. However, in real-world applications, the distributions may only be characterized subjectively; that is, failure and repair patterns are frequently summarized with everyday language descriptions of central tendencies, such as ''the mean failure rate is approximately 3 per day'' rather than with complete probability distributions. The looseness with which the system measures are reported is revealing of the uncertainty concerning these distributions. And because times to failure and times to repair are therefore possibility rather than probabilistic, the reliability (or availability) problem becomes one of decision making in the context of risk. To broaden applications of reliability and availability analysis in engineering, general science, and management [11] [12] this article extends it to fuzzy environments [13].

There are many studies on stochastic models with fuzzy environments in recent literature base. Only few among these studies focus on repairable systems with fuzzy parameter patterns using parametric nonlinear programming [14], [15]. [16], [17] study repairable series system with standby redundancy and imperfect coverage and fuzzy parameters.

Different from other studies, our model provides, a suitable estimation value form uncertain environments, a comparison discussion of using fuzzy theory and conventional method and also a comparison between parallel (active redundancy) and series system in fuzzy world when we have standby redundancy.

## II. PROPOSED METHOD

### A. Model Description

Please System reliability is generally defined as the probability that a system performs its intended function under operating conditions for a specified period of time (Meeker and Escobar1998). In practice we use mean time to failure (MTTF) to show the system reliability. On the other hands this concept changes its meaning from mean time to failure (MTTF) to mean time between failures (MTBF) for repairable systems. Also system availability is the probability that a system performs its intended function at time t under operating conditions. Reliability concept is used when the system isn't repairable and availability is for repairable system. But under certain conditions and for specific time intervals, we can use reliability concept for repairable systems.

We study a redundant repairable system with two identical operating units, which work independently and simultaneously in the parallel conditions as we call them active redundancy, and one standby. So we have both active and standby redundancy. Consider it may be impossible covered on the failure of an operating unit (or standby), even when replacing a failed unit with a standby. A detailed description of the repairable system is given in [16]. For the active and standby redundancy system that presented in this paper, we change assumption number 2 as follow:

1. *The standby unit may fail before it is put into full operation and is continuously monitored by a failure detection device. All the units both operating and standby are repairable. An operating unit fails independent of the state of the other operating unit and has an exponential time-to-failure distribution with rate parameter $\lambda$. When an operating unit fails, it is immediately replaced by the standby if it is available. The standby fails independently of the state of the operating units and has an exponential time-to-failure distribution with rate parameter $\theta$ ($0 \leq \theta \leq \lambda$).*

2. *If an operating unit fails, even active redundancy, it is immediately detected, located, and replaced with coverage probability c with the standby if it is available. It is assumed that replacement is instantaneous. We also assume that the coverage factor is the same for an operating unit failure as for a standby unit failure (both denoted c). We define the unsafe failure 1 (UF1) state of the system which two primary units are working and standby unit is valid and suddenly breakdown occurred and it can't be covered. Also unsafe failure 2 (UF2) state of the system which two units are working and the third unit (standby unit) is undergoing repair and then breakdown occurred and it can't be covered. We assume operating unit failure in the unsafe failure state is cleared by a system reboot. Reboot delays are exponentially distributed with rate parameter $\beta$. We define P0 as the state in which two units of three are undergoing repair and system operate with only one unit and then breakdown occurs so it means that the standby is emptied and system stop working. For this system as we describe, complete failure happen when one of these three states, UF1, UF2 and P0 occur.*

3. *It is assumed that when the standby replaces a failed unit and commences operation, its failure characteristics become those of an operating unit. A failed unit immediately enters the repair facility and is treated as a standby after repair. Repair time is exponentially distributed with rate parameter $\mu$.*

4. *If an operating unit or standby is undergoing repair, subsequently failed units must wait in a queue until the repair facility is available. The failed units arriving at the repair facility are served in order of arrival (FIFO system). In addition, we assume that the repair time is independent of state of the system.*

Finally, let n index the states corresponding to the normal number of units in the repairable system (i.e. n = 3, 2) and N(t) denote the state of the repairable system at time t. Then {N(t); t ≥ 0} is a continuous time Markov process with six states.

In [17] found the differential equation for only two series units with one standby. In this paper we assume that the process is initially in state 3, so that $P_3(0) = 1$, $P_2(0) = 0$, $P_1(0) = 0$, $P_{UF1}(0) = 0$ and $P_{UF2}(0) = 0$. Thus, the system differential equations using Laplace transforms are obtained in terms of $\lambda$, $\mu$, and $\theta$ as follows:

$$s\widetilde{P_3}(s) - 1 = -(2\lambda + \theta)\widetilde{P_3}(s) + \mu \widetilde{P_2}(s), \quad (1)$$

$$s\widetilde{P_2}(s) = -(2\lambda + \mu)\widetilde{P_2}(s) + c(2\lambda + \theta)\widetilde{P_3}(s) + \mu \widetilde{P_1}(s), \quad (2)$$

$$s\widetilde{P_1}(s) = -(\lambda + \mu)\widetilde{P_1}(s) + 2c\lambda \widetilde{P_2}(s), \quad (3)$$

$$s\widetilde{P_0}(s) = \lambda \widetilde{P_1}(s), \quad (4)$$

$$s\widetilde{P_{UF_1}}(s) = (1-c)(2\lambda + \theta)\widetilde{P_3}(s), \quad (5)$$

$$s\widetilde{P_{UF_2}}(s) = (1-c)(2\lambda)\widetilde{P_2}(s), \quad (6)$$

If we assume that $P_0$, $UF_1$ and $UF_2$ are system down states. Thus, the reliability function can be formalized as

$$R_Z(t) = 1 - P_0(t) - P_{UF_1}(t) - P_{UF_2}(t), \quad t \geq 0. \quad (7)$$

the Laplace transform of the failure density:

$$Z(t) = -\frac{dR_z(t)}{dt} = \frac{d(P_0(t) + P_{UF_1}(t) + P_{UF_2}(t))}{dt},$$

can then be written:

$$\widetilde{Z}(s) = s\widetilde{P_0}(0) + P_0(0) + s\widetilde{P}_{UF_1}(s) - P_{UF_1}(0) + s\widetilde{P}_{UF_2}(s) - P_{UF_2}(0).$$

### B. Fuzzy Repairable System

In this paper, we extend the applicability of the repairable system by allowing the system parameters to follow fuzzy specification. Let the failure rate of an operating unit $\lambda$, the failure rate of standby unit $\theta$, and the repair rate of failed units $\mu$ are approximately known and can be represented by the fuzzy sets $\widetilde{\lambda}$, $\widetilde{\theta}$ and $\widetilde{\mu}$, respectively. Let $\eta_{\widetilde{\lambda}}(x)$, $\eta_{\widetilde{\theta}}(v)$ and $\eta_{\widetilde{\mu}}(y)$ denote the membership functions of $\tilde{\lambda}$, $\tilde{\theta}$ and $\tilde{\mu}$, respectively. We then have the following fuzzy sets:

$$\tilde{\lambda} = \left\{ (x, \eta_{\tilde{\lambda}}(x)) \middle| x \in X \right\}, \quad (16a)$$

$$\tilde{\theta} = \left\{ (v, \eta_{\tilde{\theta}}(v)) \middle| v \in V \right\}, \quad (16b)$$

$$\tilde{\mu} = \left\{ (y, \eta_{\tilde{\mu}}(y)) \middle| y \in Y \right\}, \quad (16c)$$

where $X, V,$ and $Y$ are the crisp universal sets of the failure rate of an operating unit, the failure rate of the standby unit, and the service rate of a failed unit, respectively. Let $f(x,v,y)$ denote the system characteristic of interest (e.g., MTBF or availability) is defined as:

$$\eta_{\tilde{T}}(z) = \sup_{\Omega_1} \min\left\{ (\eta_{\tilde{\lambda}}(x), \eta_{\tilde{\theta}}(v), \eta_{\tilde{\mu}}(y)) \middle| z = Q \right\}. \quad (18a)$$

and

$$\eta_{\tilde{A}}(z) = \sup_{\Omega_2} \min\left\{ (\eta_{\tilde{\lambda}}(x), \eta_{\tilde{\theta}}(v), \eta_{\tilde{\mu}}(y), \eta_{\tilde{\beta}}(w)) \middle| z = R \right\}. \quad (18b)$$

with

$\Omega_1 = \{x \in X, v \in V, y \in Y | x > 0, v > 0, y > 0\}$ and
$\Omega_2 = \{x \in X, v \in V, y \in Y, w \in W | x > 0, v > 0, y > 0, w > 0\}.$

The membership functions in (18) are not in the usual forms for practical use making it very difficult to imagine their shapes.

## III. RESULT & DISCUSSION

### A. Numerical Example

We consider an example motivated by a real-life system to demonstrate the practical use of the proposed solution that also used and described in [16]. An electric power plant has two main coal power generators that operate in parallel system and one light-diesel turbo generator for standby use. The capacity of each coal power generator is 300 MW and the capacity of the light-diesel generator is 150 MW. The standby generator may fail while waiting to be put into full operation and it is continuously monitored by a failure detection device. Operating and standby generators are repairable. If an operating generator fails, it is immediately detected, located, and replaced with a coverage probability c with a standby if one is available. It is assumed that replacement is instantaneous [18]. For efficiency, the management wants to get the system characteristics such as MTBF and availability.

### B. Summary of Results

One advantage of using the fuzzy theory to analyze the system characteristics of a repairable system at different possibility α levels is that the manager can adaptively adjust the components of the system based on the damaged level of the system affected by the broken components. For example, the manager may consider the costs related to a repairable system to decide the proper range of the availability from 0.9126 to 0.9558. From Table I, the intervals of the failure rates of an operating unit, the failure rates of a standby unit, the repair rates of a failed unit, and the reboot rates are, respectively, and corresponded α level is 0.9. There are two situations for the manger to decide the suitable maintain strategy.

First, when the broken component (the coal power generator) will cause a seriously damage of the system, the manager can make a conservative maintaining plan. The coal power generator can be replaced or fixed (maintained) when it has been used for 0.59 time unit. In addition, if the range of the repair time is controlled to be [5.9, 7.1] time unit, the manager can get the required availability of this system. Second, when the broken component (the coal power generator) will not cause a seriously damage of the system, the manager can make a cost-effective maintaining plan. The component will be replaced or fixed (maintained) only when it has been used nearly 0.71 time unit. The required availability is acquired when the range of the repair time is also controlled to be [5.9, 7.1] time unit. On the other hand, the traditional approach cannot provide this useful information. Note that when we analyze the system characteristics under the same conditions, we can find the MTBF (or availability) of parallel configuration is larger than the value of series configuration.

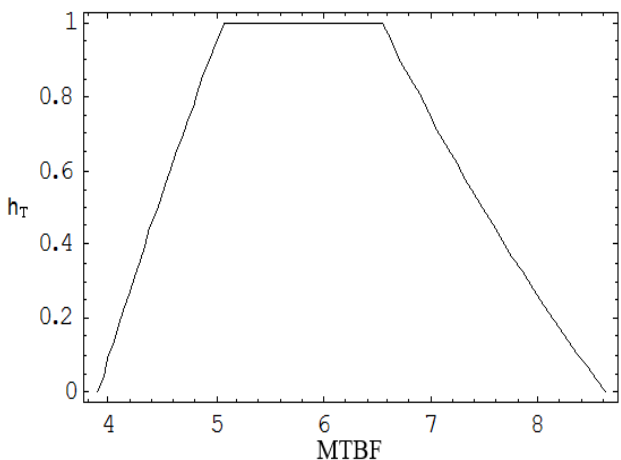

Fig. 1. The membership function for fuzzy MTBF.

TABLE I
A-CUTS OF THE FAILURE RATE OF AN OPERATING UNIT.

| α | $x_\alpha^L$ | $x_\alpha^U$ | $v_\alpha^L$ | $v_\alpha^U$ | $y_\alpha^L$ | $y_\alpha^U$ | $(T)_\alpha^L$ | $(T)_\alpha^U$ |
|---|---|---|---|---|---|---|---|---|
| 0.00 | 0.50 | 0.80 | 0.10 | 0.40 | 3.00 | 6.00 | 3.8952 | 8.6229 |
| 0.10 | 0.51 | 0.79 | 0.11 | 0.39 | 3.10 | 5.90 | 4.0030 | 8.3722 |
| 0.20 | 0.52 | 0.78 | 0.12 | 0.38 | 3.20 | 5.80 | 4.1126 | 8.1330 |
| 0.30 | 0.53 | 0.77 | 0.13 | 0.37 | 3.30 | 5.70 | 4.2242 | 7.9046 |
| 0.40 | 0.54 | 0.76 | 0.14 | 0.36 | 3.40 | 5.60 | 4.3377 | 7.6859 |
| 0.50 | 0.55 | 0.75 | 0.15 | 0.35 | 3.50 | 5.50 | 4.4534 | 7.4764 |
| 0.60 | 0.56 | 0.74 | 0.16 | 0.34 | 3.60 | 5.40 | 4.5712 | 7.2752 |
| 0.70 | 0.57 | 0.73 | 0.17 | 0.33 | 3.70 | 5.30 | 4.6913 | 7.0817 |
| 0.80 | 0.58 | 0.72 | 0.18 | 0.32 | 3.80 | 5.20 | 4.8139 | 6.8955 |
| 0.90 | 0.59 | 0.71 | 0.19 | 0.31 | 3.90 | 5.10 | 4.9390 | 6.7159 |
| 1.00 | 0.60 | 0.70 | 0.20 | 0.30 | 4.00 | 5.00 | 5.0669 | 6.5424 |

*C. Discussion*

From this example, more information is provided to the manager and the manufacturer. For example, a manager may consider the costs related to a repairable system to decide the optimal maintenance strategy; to do so, he or she can set the range of MTBF to be between 4.939 and 6.716 h to reflect the desired repair rate and find that the corresponding α level is 0.9. In other words, the manager can determine that the repair rate should be between 3.9 and 5.1 h. Similarly, a manager can set the range of availability to be in the interval [0.9126, 0.9558] to reflect the desired repair rate and find that the corresponding α level.

## IV. CONCLUSION

This paper applies the concepts of a-cuts and Zadeh's extension principle to a repairable system with two primary units in parallel which are active and one standby redundancy in the context of imperfect coverage and constructs membership functions of MTBF and availability using paired NLP models. Following the proposed approach, α-cuts of the membership functions are found and their corresponding interval limits inverted to attain explicit closed-form expressions for the system characteristics. We illustrate the validity of the proposed approach by an example from [19]. Note that when the membership function intervals cannot be inverted explicitly, system management or designers can specify the system characteristics of interest, perform numerical calculations, examine the corresponding α-cuts, and use this information to develop or improve system processes.

Although the proposed procedure in this paper is for a one-unit repairable system with active and standby redundancy, it can easily be applied to the generalized case with M primary units and S standbys. The procedure developed is based on the case with the exponential time to failure and the exponential time to repair. When distribution of time to failure or time to repair is not exponential, it still can be implemented if the system characteristics are obtained explicitly in terms of parameters. Moreover, the coverage factor for an operating unit failure may not be the same as a standby unit failure. Also time to repair can be dependent on the number of units that waiting in the queuing system. Although the primary application of proposed model is in reliability and warranty related decision making however it can also in other domain such as healthcare [20] surgical units [21], and robotic surgery devices [22], [23].

## REFERENCES

[1] M. J. Fard, S. Ameri, and A. Z. Hamadani, "Bayesian Approach for Early Stage Reliability Prediction of Evolutionary Products," in *International Conference on Operations Excellence and Service Engineering (IEOM)*, 2015, pp. 1–11.
[2] H. Pham, "Reliability analysis of a high voltage system with dependent failures and imperfect coverage," *Reliab. Eng. Syst. Saf.*, vol. 37, no. 1, pp. 25–28, Jan. 1992.
[3] K. S. Trivedi, *Probability and statistics with reliability, queuing, and computer science applications*. Wiley, 2002.
[4] A. T. de Almeida and F. M. Campello de Souza, "Decision theory in maintenance strategy for a 2-unit redundant standby system," *IEEE Trans. Reliab.*, vol. 42, no. 3, pp. 401–407, 1993.
[5] M. Gururajan and B. Srinivasan, "A complex two-unit system with random breakdown of repair facility," *Microelectron. Reliab.*, vol. 35, no. 2, pp. 299–302, Feb. 1995.
[6] R. Subramanian and V. Anantharaman, "Reliability analysis of a complex standby redundant systems," *Reliab. Eng. Syst. Saf.*, vol. 48, no. 1, pp. 57–70, Jan. 1995.
[7] S. P. Rajamanickam and B. Chandrasekar, "Reliability measures for two-unit systems with a dependent structure for failure and repair times," *Microelectron. Reliab.*, vol. 37, no. 5, pp. 829–833, May 1997.
[8] R. Billinton and J. Pan, "Optimal maintenance scheduling in a two identical component parallel redundant system," *Reliab. Eng. Syst. Saf.*, vol. 59, no. 3, pp. 309–316, Mar. 1998.
[9] J. H. Seo, J. S. Jang, and D. S. Bai, "Lifetime and reliability estimation of repairable redundant system subject to periodic alternation," *Reliab. Eng. Syst. Saf.*, vol. 80, no. 2, pp. 197–204, 2003.
[10] M. Jain and R. Gupta, "Optimal replacement policy for a repairable system with multiple vacations and imperfect fault coverage," *Comput. Ind. Eng.*, vol. 66, no. 4, pp. 710–719, 2013.
[11] S. Ameri, M. J. Fard, R. B. Chinnam, and C. K. Reddy, "Survival Analysis based Framework for Early Prediction of Student Dropouts," in *Proceedings of the 25th ACM International on Conference on Information and Knowledge Management - CIKM '16*, 2016, pp. 903–912.
[12] M. J. Fard, S. Chawla, and C. K. Reddy, "Early-Stage Event Prediction for Longitudinal Data," in *Pacific-Asia Conference on Knowledge Discovery and Data Mining*, 2016, pp. 139–151.
[13] L. A. Zadeh, "Fuzzy Sets as a Basic for aTheory of Possibility," *Fuzzy Sets Syst.*, vol. 1, no. 1, pp. 3–28, 1978.
[14] J.-C. Ke, H.-I. Huang, and C.-H. Lin, "Fuzzy analysis for steady-state availability: a mathematical programming approach," *Eng. Optim.*, vol. 38, no. 8, pp. 909–921, 2007.
[15] H.-I. Huang, C.-H. Lin, and J.-C. Ke, "Parametric nonlinear programming approach for a repairable system with switching failure and fuzzy parameters," *Appl. Math. Comput.*, vol. 183, no. 1, pp. 508–517, 2006.
[16] J.-C. Ke, H.-I. Huang, and C.-H. Lin, "A redundant repairable system with imperfect coverage and fuzzy parameters," *Appl. Math. Model.*, vol. 32, no. 12, pp. 2839–2850, 2008.
[17] M. J. Fard, S. Ameri, S. R. Hejazi, and A. Zeinal Hamadani, "One-unit repairable systems with active and standby redundancy and fuzzy parameters," *Int. J. Qual. Reliab. Manag.*, vol. 34, no. 3, pp. 446–458, Mar. 2017.
[18] M. J. Fard, P. Wang, S. Chawla, and C. K. Reddy, "A Bayesian Perspective on Early Stage Event Prediction in Longitudinal Data," *IEEE Trans. Knowl. Data Eng.*, vol. 28, no. 12, pp. 3126–3139, Dec. 2016.
[19] K.-H. Wang, Y.-C. Liu, and W. L. Pearn, "Cost benefit analysis of series systems with warm standby components and general repair time," *Math. Methods Oper. Res.*, vol. 61, no. 2, pp. 329–343, Jun. 2005.
[20] J. F.-F. Yao and J.-S. Yao, "Fuzzy decision making for medical diagnosis based on fuzzy number and compositional rule of inference," *Fuzzy Sets Syst.*, vol. 120, no. 2, pp. 351–366, Jun. 2001.
[21] M. J. Fard, "Computational Modeling Approaches for Task Analysis in Robotic-Assisted Surgery," Wayne State University, 2016.
[22] H.-C. Liu, J. Wu, and P. Li, "Assessment of health-care waste disposal methods using a VIKOR-based fuzzy multi-criteria decision making method," *Waste Manag.*, vol. 33, no. 12, pp. 2744–2751, Dec. 2013.
[23] M. J. Fard, S. Ameri, R. B. Chinnam, and R. D. Ellis, "Soft Boundary Approach for Unsupervised Gesture Segmentation in Robotic-Assisted Surgery," *IEEE Robot. Autom. Lett.*, vol. 2, no. 1, pp. 171–178, Jan. 2017.